\makeatletter\def\graphicscache@inhibit{true}\makeatother
\documentclass{llncs}
\usepackage[utf8]{inputenc}
\usepackage{url}
\usepackage[hidelinks]{hyperref}
\usepackage{graphicx}
\usepackage{setspace}
\usepackage[T1]{fontenc}
\usepackage{multicol}
\usepackage{amsmath,amssymb}
\usepackage{epstopdf}
\usepackage{graphicx,color}
\usepackage{xspace}
\usepackage{pifont}
\usepackage{balance}
\usepackage[per-mode=symbol,binary-units=true,range-units=single,range-phrase=-]{siunitx}
\usepackage{textcomp} %
\usepackage{booktabs}
\usepackage{paralist}
\usepackage{footmisc}
\usepackage{comment}
\usepackage{subcaption}
\usepackage{layouts}
\usepackage{cancel}
\usepackage{url}
\usepackage{bm}
\usepackage[percent]{overpic}
\usepackage{tabularx} %
\usepackage{multirow} %
\usepackage{makecell} %
\usepackage{tikz}
\usepackage{pgfplots}
\usepackage{gnuplot-lua-tikz}
\usepackage{pgfplots}
\pgfplotsset{compat=1.15}
\usepgfplotslibrary{groupplots}
\usetikzlibrary{fit,shapes.callouts,shapes.geometric,backgrounds,positioning,arrows.meta}
\graphicspath{{./figures/}}
\usepackage[linesnumbered,ruled,vlined]{algorithm2e}
\SetKwInput{KwInput}{Input}                %
\SetKwInput{KwOutput}{Output}              %

\newcommand{\ie}{i.e.,\ }
\newcommand{\eg}{e.g.,\ }

\newcommand{\etal}{\xspace{}et al.\xspace}

\newcommand{\reffig}[1]{Fig.~\ref{#1}}
\newcommand{\reftab}[1]{Tab.~\ref{#1}}
\newcommand{\refsec}[1]{Sec.~\ref{#1}}
\newcommand{\refalg}[1]{Algorithm~\ref{#1}}
\newcommand{\refeq}[1]{(\ref{#1})}

\usepackage{graphicscache}

\begin{document}

\frontmatter          %

\pagestyle{headings}  %
\addtocmark{Two-step Planning of Dynamic UAV Trajectories using Iterative $\delta$-Spaces} %

\mainmatter           %

\title{Two-step Planning of Dynamic UAV Trajectories using Iterative $\delta$-Spaces}
\titlerunning{$\delta$-Space Trajectory Planning}  %
\author{Sebastian Schr\"{a}der \and Daniel Schleich \and Sven Behnke}
\authorrunning{Schr\"{a}der, Schleich, and Behnke} %
\institute{Institute for Computer Science VI, Autonomous Intelligent Systems,\\
University of Bonn, Germany,\\
\email{schleich@ais.uni-bonn.de}}

\maketitle            %

\begin{tikzpicture}[remember picture,overlay]
\node[anchor=north,align=center,font=\sffamily] at (current page.north) {%
  \footnotesize \textbf{Accepted final version.} 17th International Conference on Intelligent Autonomous Systems (IAS), Zagreb, Croatia, to appear June 2022. 
};
\end{tikzpicture}%

\begin{abstract}
UAV trajectory planning is often done in a two-step approach, where a low-dimensional path is refined to a dynamic trajectory. The resulting trajectories are only locally optimal, however.
On the other hand, direct planning in higher-dimensional state spaces generates globally optimal solutions but is time-consuming and thus infeasible for time-constrained applications.
To address this issue, we propose $\delta$-Spaces, a pruned high-dimensional state space representation for trajectory refinement. 
It does not only contain the area around a single lower-dimensional path but consists of the union of multiple near-optimal paths.
Thus, it is less prone to local minima.
Furthermore, we propose an anytime algorithm using $\delta$-Spaces of increasing sizes.

We compare our method against state-of-the-art search-based trajectory planning methods and evaluate it in 2D and 3D environments to generate second-order and third-order UAV trajectories.
\end{abstract}

\section{Introduction}
\label{sec:Introduction}
\vspace{-0.2cm}
\begin{figure}[t]
    \centering
    \framebox{\includegraphics[width=0.7\linewidth]{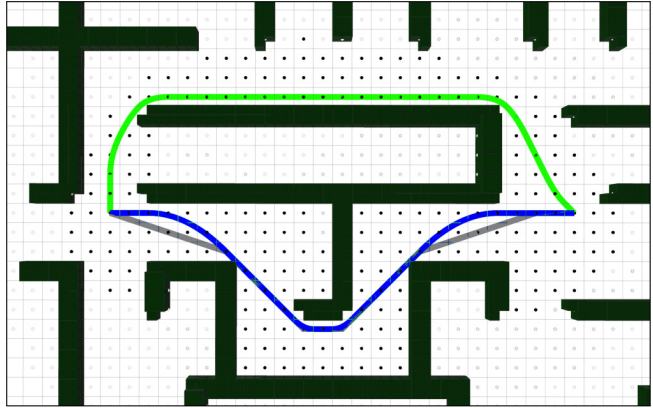}}
    \caption{Refining a 2D path (grey) to a dynamic trajectory. The blue trajectory is obtained by planning in a tunnel around the 2D path. The green trajectory is obtained by planning in the $\delta$-Space (black dots). Although the green trajectory is more distant to the shortest 2D path, it can be tracked at higher velocities and has thus a shorter flight time.}
    \label{fig:teaser}
\end{figure} 

In recent years, unmanned aerial vehicles (UAVs) have gained increasing interest for practical usage in many different fields like industrial inspection, agriculture, search \& rescue, and delivery.
Due to their ability for agile flight and high velocities, UAVs have huge potential for time-critical applications, \eg in disaster-response scenarios.
When flying close to obstacles, the systems' dynamics have to be considered during trajectory planning.

Liu \etal~\cite{liu_2017_iros} directly incorporate UAV dynamics into a global planner.
They use high-dimensional state lattices, from which they extract the optimal trajectory using A* and a heuristic based on the solution of a Linear Quadratic Minimum Time problem.
Such approaches become infeasible for large environments, however, since searching high-dimensional state spaces is computationally expensive.
Thus, a reduction of the state space size is necessary, either explicitly via pruning or implicitly using more informed search heuristics \cite{liu_2018_icra}.

Many existing methods reduce the computational load by splitting the trajectory planning problem into two parts.
A lower-dimensional geometric path can efficiently be found using search-based \cite{hart_1972_sigart} or sampling-based \cite{lavalle_1998} planners.
After choosing an initial time allocation, this path can be refined to a high-dimensional dynamic trajectory \cite{nieuwenhuisen_2016_iros}.
To optimize the smoothness of high-dimensional trajectories, multiple methods haven been proposed, including gradient-based optimizers \cite{kalakrishnan_2011_icra,zucker_2013_sage}, B-splines \cite{arney_2007_iciafs}, and neural networks \cite{simon_1993_ras}.
These approaches only locally optimize the trajectory and are prone to local minima, however.
The shortest geometric path towards the goal is not necessarily a good initialization for trajectory optimization, especially when it contains many turns or if we consider non-static initial states.
A longer path might suit the UAV dynamics better, allowing higher velocities and thus resulting in shorter flight times.

In this work, we propose a method where the high-dimensional trajectory generation is not guided by a single shortest path but multiple near-optimal paths are considered.
For this, we introduce the $\delta$-Space, a set of paths whose length is within a certain sub-optimality bound $\delta$.
The motivation behind this definition is visualized in \reffig{fig:teaser}.
By allowing larger deviations from the shortest low-dimensional path, we find trajectories that can be tracked at higher velocities and reduce the risk of local minima.
Additionally, our method considers the structure of the environment and can efficiently prune dead ends, which is not possible if a tunnel with large radius around a single low-dimensional path is used.

In summary, the main contributions of this paper are:
\begin{itemize}
 \item the introduction of $\delta$-Spaces to prune high-dimensional state spaces while reducing the risk of local minima,
 \item an anytime algorithm allowing to iteratively increase the size of the $\delta$-Space,
 \item and the application to second-order and third-order dynamic trajectory planning for UAVs, including an efficient search heuristic that reuses information from the $\delta$-Space construction.
\end{itemize}

\section{Related Work}
\label{sec:Related_Work}
\vspace{-0.2cm}
Several approaches to reduce the size of high-dimensional state spaces have been proposed.
A common method is to restrict the high-dimensional search to the vicinity of the solution of a lower-dimensional problem representation.
For example, this can be done by only considering high-dimensional states, whose position components lie within a tunnel around a previously computed geometric path \cite{stachniss_2002_iros}.
Such approaches are prone to local minima, however.
The high-dimensional search might not even find a solution due to kinodynamic constraints if the tunnel size is too small.

Instead of pruning the state space, the solution of a lower-dimensional problem formulation can be used as a heuristic to guide the high-dimensional search.
MacAllister \etal~\cite{macallister_2013_icra} solve a simplified spatial path-planning problem using breadth-first search to guide the planning, but they only consider 4D state lattices.
Liu \etal~\cite{liu_2018_icra} proposed to iteratively compute first-, second- and third-order UAV trajectories, where the lower-dimensional solution is used to guide the next-higher dimensional trajectory.
This ensures that a solution can be found since large deviations from the low-dimensional trajectory are discouraged but possible.
The resulting trajectories are no longer globally optimal, however.
One approach to address the local minima problem of hierarchical methods was introduced by Ding \etal~\cite{ding_2019_tro}, who proposed a global B-spline-based kinodynamic search algorithm.

Another approach is to reduce the state space size globally using multiresolution.
Du \etal~\cite{du_2020_socs} perform multiple A* searches in parallel, each at a different resolution of the state space.
This approach has been extended to an anytime algorithm by Saxena \etal~\cite{saxena_2021_arxiv}.
The idea of multiresolution has also be considered in combination with high-dimensional state lattices.
Gonz\'alez-Sieira \etal~\cite{gonzalez_2019_icra} adapt the resolution of a state lattice locally to the environment.
This is done by grouping motion primitives that represent similar actions and considering only the longest collision-free primitive from each group.
Schleich and Behnke~\cite{schleich_2021_icra} apply the idea of local multiresolution \cite{behnke_2003} to state lattices for faster replanning of dynamic UAV trajectories.

Besides the resolution, the state dimensionality can also be adapted to accelerate the planning.
Gochev \etal~\cite{gochev_2011_socs} represent the state space locally at different dimensionalities.
This allows them to only use high-dimensional state representations when necessary, \ie in narrow areas of the environment, and to plan in a lower-dimensional space otherwise.
Gochev \etal~\cite{gochev_2013_icaps} extend this method with an incremental version of weighted A* to achieve a better performance.
Vemula \etal~\cite{vemula_2016_socs} apply adaptive dimensionality to the problem setting of path planning with dynamic obstacles.

In this work, we propose a state space pruning technique similar to the tunnel method \cite{stachniss_2002_iros} but use multiple lower-dimensional paths to determine which high-dimensional states are pruned.
This makes our method less prone to local minima.
Additionally, we can iteratively increase the size of our state space efficiently in an anytime fashion to further improve the solution quality.
 
\section{Method}
\label{sec:Method}
\vspace{-0.2cm}
Applying search-based planning methods to high-dimensional state spaces is computationally expensive. 
Thus, many methods reduce the complexity of such search problems by projecting the high-dimensional robot state onto a lower-dimensional space, where a solution can be found with less effort. 
This low-dimensional solution can then be used to accelerate high-dimensional planning, either by guiding the search or by pruning the state space. 
In the context of UAV trajectory planning, the first step usually consists of finding an optimal spatial path without dynamic consideration. 
Here, optimality is usually defined as minimizing the path length.
However, if the shortest path contains many turns, it can only be tracked by the UAV at low velocities.
A longer path might suit the UAV dynamics much better and can lead to shorter flight times.
Thus, the high-dimensional trajectory generation should not be guided by a single shortest path.
Instead, we propose to use the $\delta$-Space, \ie the union of all paths whose length is within a sub-optimality bound $\delta$.
The high-dimensional trajectory search can then be restricted to states whose position components are part of this space.
In \refsec{sec:delta_space}, we give a formal definition of the $\delta$-Space and how it is used to solve high-dimensional search-problems.

The choice of $\delta$ significantly influences the performance of the proposed method.
A small value results in fast planning times but yields only locally optimal solutions.
When choosing high values, the search is less prone to local minima, but planning times are increased.
In \refsec{sec:iterative_delta}, we propose an efficient method how to iteratively increase $\delta$ while reusing the previous search results.
Thus, an initial result can be found quickly using a small value of $\delta$ and the obtained solution can be subsequently improved in an anytime fashion.

Finally, we point out that $\delta$-Spaces cannot only be used to prune high-dimensional state spaces.
When generating the $\delta$-Space, we already explore the structure of the environment.
The obtained information can be integrated into a search heuristic to guide the higher-dimensional planning.
Since this depends on the specific problem setting, we present an example for such a heuristic in \refsec{sec:delta_application}, which describes the application of $\delta$-Spaces to dynamic UAV trajectory planning.

\subsection{$\delta$-Space Definition}
\label{sec:delta_space}

We consider a planning problem with a set of high-dimensional robot states $\mathcal S_h$, valid state transitions $\mathcal E_h \subset \mathcal S_h \times \mathcal S_h$, a cost function $c_h: \mathcal E_h \rightarrow \mathbb R_{\geq 0}$, an initial state $s$, and a goal state $g$.
A trajectory $\mathcal T_{h, s \mapsto g}$ from $s$ to $g$ is a sequence $(s_0, \dots, s_n)$ with $(s_i, s_{i+1}) \in \mathcal E_h$, $s_0=s$, and $s_n=g$.
The cost of a trajectory is defined as the sum of the individual state transition costs, \ie
\begin{equation}
\mathcal C_h(\mathcal T_{h, s \mapsto g}) := \sum\limits_{i=0}^{n-1} c_h(s_i, s_{i+1}).
\label{eq:trajectory_cost}
\end{equation}
We are interested in finding a trajectory $\mathcal T^{opt}_{h, s \mapsto g}$ minimizing these costs.
In the following, we use the shorthand notation $\mathcal C_h( s, g ) := \mathcal C_h( \mathcal T^{opt}_{h, s \mapsto g} ) $ for the costs of an optimal trajectory from $s$ to $g$.

To support the planning, we additionally define an abstract, lower-di\-men\-sion\-al problem representation, consisting of states $\mathcal S_l$, transitions $\mathcal E_l \subset \mathcal S_l \times \mathcal S_l$ and costs $\mathcal C_l: \mathcal E_l \rightarrow \mathbb R$.
A high-dimensional robot state can be transformed into an abstract one using the projection $\pi:\mathcal S_h \rightarrow \mathcal S_l$.

For a given instance of the planning task, we define the $\delta$-Space $\mathcal S_l^\delta$ as 
\begin{equation}
\mathcal S_l^\delta := \left\{s_l\in\mathcal S_l \; | \; \mathcal C_l( \pi(s), s_l) + \mathcal C_l( s_l, \pi(g) ) \leq \mathcal C_l( \pi(s), \pi(g) ) + \delta \right\}.
 \label{eq:delta_space}
\end{equation}
Note that this definition depends on the specific start and goal states of the current planning instance.

For solving the higher-dimensional planning problem, we restrict the state space to states for which the corresponding abstract state is part of the $\delta$-Space.
Thus, we consider the pruned state space
\begin{equation}
\mathcal S_h^\delta := \left\{s_h\in\mathcal S_h \; | \; \pi(s_h) \in \mathcal S_l^\delta \right\}.
 \label{eq:pruned_state_space}
\end{equation}
Additionally, we define the closure $\overline{\mathcal S}_h^\delta$ of this set as the set of all states that can be reached from $\mathcal S_h^\delta$, \ie
\begin{equation}
\overline{\mathcal S}_h^\delta := \left\{s_1 \in \mathcal S_h \; | \; \exists s_0 \in \mathcal S_h^\delta: (s_0,s_1)\in\mathcal E_h \right\}.
 \label{eq:delta_space_closure}
\end{equation}
Correspondingly, we define the set of possible state transitions
\begin{equation}
\mathcal E_h^\delta := \left\{e_h:=(s_0,s_1)\in\mathcal E_h \; | \; s_0, s_1 \in \mathcal S_h^\delta \right\}
 \label{eq:pruned_transitions}
\end{equation}
and
\begin{equation}
\overline{\mathcal E}_h^\delta := \left\{e_h:=(s_0,s_1)\in\mathcal E_h \; | \; s_0, s_1 \in \overline{\mathcal S}_h^\delta \right\}.
 \label{eq:closure_transitions}
\end{equation}
To ensure that only states from $\mathcal S_h^\delta$ are considered during planning, we adjust the cost function to be
\begin{equation}
    c_h^\delta(e_h)= 
    \begin{cases}
        c_h(e_h),& \text{if } e_h \in \mathcal E_h^\delta \\
        \infty,              & \text{otherwise.}
    \end{cases}
 \label{eq:delta_cost_function}
\end{equation}
States from the boundary of the state space, \ie from $\overline{\mathcal S}_h^\delta \setminus \mathcal S_h^\delta$, are only needed for the anytime planning algorithm in \refsec{sec:iterative_delta} to determine possible states that will be added to the search space when increasing $\delta$.

Although the definition of the $\delta$-Space can be applied to continuous state spaces as well, we restrict ourselves in this work to discretized states.
Thus, we can cast planning as graph search.
After the $\delta$-Space is constructed, the high-dimensional trajectory can be generated by using graph search algorithms like A* on the state lattice graph $\mathcal G^\delta_h( \overline{\mathcal S}_h^\delta, \overline{\mathcal E}_h^\delta )$ consisting of nodes $\overline{\mathcal S}_h^\delta$ and edges $\overline{\mathcal E}_h^\delta$.

 \begin{figure}[t]
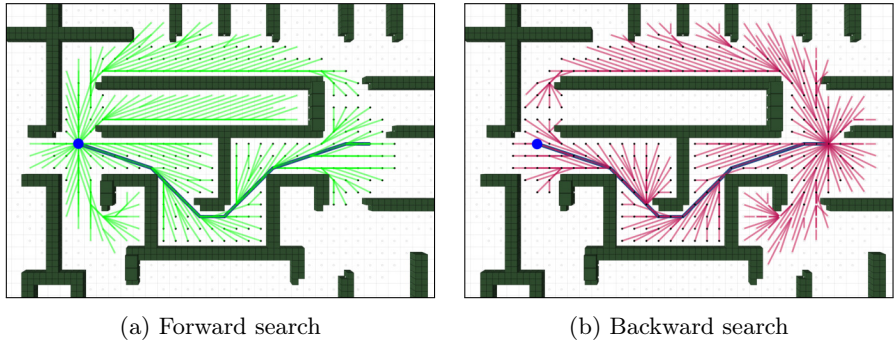

    \begin{subfigure}{.49\textwidth}
    \centering
    \framebox{\includegraphics[width=.95\textwidth]{img/delta_space_forwards.png}}
    \captionsetup{justification=centering}
    \caption{Forward search}
    \end{subfigure}
    \begin{subfigure}{.49\textwidth}
    \centering
    \framebox{\includegraphics[width=.95\textwidth]{img/delta_space_backwards.png}}
    \captionsetup{justification=centering}
    \caption{Backward search} 
    \end{subfigure}
    \caption[Lower-dimensional search forwards and backwards]{Lower-dimensional A* searches to calculate the $\delta$-Space. The start state is marked with the blue circle. The nodes forming the $\delta$-Space (black dots) are expanded by both forward and backward search. The blue line is thereby the shortest path.}
    \label{fig:delta_space_generation}
\end{figure}

To construct the $\delta$-Space, we solve the lower-dimensional planning problem using A* search.
According to \refeq{eq:delta_space}, for each state $s_l \in \mathcal S_l$, we need to know both the cost from the start state to $s_l$ as well as the cost from $s_l$ towards the goal.
Thus, we use two searches, one forward search from the start and one backward search starting at the goal, as shown in \reffig{fig:delta_space_generation}.
These searches are independent of each other and can be parallelized for a better performance.
If a state $s_l$ has been expanded by both searches, we can easily decide whether it belongs to the $\delta$-Space by checking the criterion \refeq{eq:delta_space}.
To find all states of the $\delta$-Space, we slightly adjust the A* algorithm:
The $f$-value of a state $s'$ is defined as the sum of the cost from start to $s'$ and an heuristic value $h(s')$ estimating the remaining cost-to-go. 
If the heuristic function $h$ used for the A* search is admissible, $f(s')$ represents a lower threshold on the costs of an optimal trajectory containing $s'$.
Thus, after finding the optimal trajectory $\mathcal T^{opt}_{l, \pi(s) \mapsto \pi(g)}$, we extract its costs and continue to expand states until the next state to be expanded has an $f$-value exceeding $\mathcal C_l(\mathcal T^{opt}_{l, s \mapsto g}) + \delta$.
For an admissible heuristic, we thus can ensure that all states of the $\delta$-Space are expanded by the forward and backward searches.

\subsection{Iterative $\delta$-Spaces}
\label{sec:iterative_delta}

In general, a small $\delta$ results in low planning times.
However, larger values of $\delta$ are necessary to avoid local minima.
In this section, we describe an anytime planning algorithm for the case where a planning time limit is set, \eg when replanning at a fixed frequency.
The anytime planner starts with a small $\delta$ to quickly generate a first solution and afterwards iteratively increases $\delta$ to further improve the solution until the planning time limit is reached.

As described above, the states within the $\delta$-Space are found by executing forward and backward A* searches until $f$-values are reached that exceed $\mathcal C_l(\mathcal T^{opt}_{l, s \mapsto g}) + \delta$.
For higher values of $\delta$, the same searches are executed, but up to higher $f$-values.
Since the $f$-values increase monotonically during A* searches, we can pause and later resume the searches for planning in $\delta$-Spaces of increasing sizes.

\begin{algorithm}
\DontPrintSemicolon
 \KwInput{Start state $s$, goal state $g$, $\delta_0$, step size $\Delta\delta$}
 \KwOutput{Trajectory $\mathcal T^{opt}_{\delta, s \mapsto g}$}
$\mathcal S_l^{\delta_0} \leftarrow $ GenerateDeltaSpace($\pi(s), \pi(g), \delta_0$)\;
\tcc{get optimal Trajectory and final search state}
$\mathcal T^{opt}_{{\delta}, s \mapsto g}$, $(\mathcal O_0, \mathcal O'_0, \mathcal F_0, \mathcal G_0) \leftarrow $ HighDimSearch($s, g, \mathcal G^{\delta_0}_h$) \;
$i \leftarrow 0$\;

\While{There is time to improve the trajectory} {
    $\delta_{i+1} \leftarrow \delta_i + \Delta\delta$\;
    $\mathcal S_l^{\delta_{i+1}} \leftarrow $ UpdateDeltaSpace($\mathcal S_l^{\delta_i}, \delta_{i+1}$)\;
    Update $\mathcal F_i, \mathcal G_i$ for all nodes in $\mathcal O'_i \cap \mathcal S_h^{\delta_{i+1}}$\;
    $\mathcal O_i \leftarrow \mathcal O_i \cup ( \mathcal O'_i \cap \mathcal S_h^{\delta_{i+1}} )$ \;
    $\mathcal T^{opt}_{{\delta}, s \mapsto g}$, $(\mathcal O_{i+1}, \mathcal O'_{i+1}, \mathcal F_{i+1}, \mathcal G_{i+1}) \leftarrow $ HighDimSearch($s, g, \mathcal G^{\delta_{i+1}}_h, (\mathcal O_i, \mathcal F_i, \mathcal G_i)$) \;
    $i \leftarrow i+1$\;
 }
 \Return $\mathcal T^{opt}_{{\delta}, s \mapsto g}$ \;
 \caption{Anytime Planning In Iterative $\delta$-Spaces}
 \label{alg:anytime_delta}
\end{algorithm}

The complete anytime planning algorithm is given in \refalg{alg:anytime_delta}.
First, we calculate the $\delta$-Space for the initial size $\delta_0$ (Line 1) and plan a high-dimensional trajectory by searching the pruned state-lattice graph $\mathcal G^{\delta_0}_h$ using A* (Line 2).
Here, we save the final state of the A* search, including the open list $\mathcal O_0$, as well as the set of $f$-values $\mathcal F_0$ and $g$-values $\mathcal G_0$.
Additionally, we keep track of all states where the search hits the boundary of the $\delta$-Space, \ie all states within $\overline{\mathcal S}_h^{\delta_0} \setminus \mathcal S_h^{\delta_0}$ that are neighbours of expanded states.
Since these states can only be reached via edges with infinite costs, they are not considered as valid states in the current search.
However, they might become valid once $\delta$ is increased.
Thus, instead of adding them to the open list,  we add them to a separate list $\mathcal O'_0$ for later use.

If the planning time limit is not met yet, we can iteratively plan in larger $\delta$-Spaces to further improve the trajectory (Line 4).
First, we increase the size of the $\delta$-Space by resuming the low-dimensional searches with increased $\delta$ (Line 5 and 6) as described above.
Afterwards, we continue the high-dimensional search on the extended state space.
Here, we first update the cost function for all states from $\mathcal O'_i$ that have been added to the larger $\delta$-Space $\mathcal S_h^{\delta_{i+1}}$ (Line 7) and add them to the open list of the previous search (Line 8).
Then, we execute the high-dimensional search on the extended search space (Line 9), starting with the already initialized open list, $f$-values and $g$-values from the previous search.
Note, that we do not include the closed list.
If the optimal trajectory contains states that were not part of the previous state space $\mathcal S_h^{\delta_i}$, the $f$-values of previously expanded states might be suboptimal.
Thus, we need to allow one re-expansion per state in each iteration.

\subsection{Application to UAV Trajectory Planning}
\label{sec:delta_application}

We apply $\delta$-Spaces to plan second-order and third-order UAV trajectories.
Thus, we model the UAV state as 6-tuples $s=(\mathbf p, \mathbf v) \in \mathbb R^6$ or 9-tuples $s=(\mathbf p, \mathbf v, \mathbf a) \in \mathbb R^9$, consisting of a discretized 3D position $\mathbf p$, velocity $\mathbf v$, and acceleration $\mathbf a$.
To compare against the State of the Art, we combine $\delta$-Spaces with the search-based trajectory generation methods from \cite{liu_2018_icra} and \cite{schleich_2021_icra}.
In both works, the set of state transitions $\mathcal E_h$ consists of motion primitives $e_{\mathbf u, \tau}$ which are generated by applying constant acceleration or jerk commands $\mathbf u$ over a short time interval $\tau$.
The corresponding costs are defined as the weighted sum of control effort and primitive duration:
\begin{equation}
c_h( e_{\mathbf u, \tau} )= ||\mathbf u||^2_2 \tau + \rho\tau.
 \label{eq:primitive_cost}
\end{equation}

For the low-dimensional problem representation, we project a high-di\-men\-sion\-al state $s=(\mathbf p, \mathbf v, \mathbf a)$ onto its position components $\mathbf p$ and use a simple 3D grid with a 26-connected neighbourhood as low-dimensional planning space.

When using search-based methods like A*, the choice of a good heuristic function is crucial for the planning performance.
In this work, we propose a heuristic that efficiently uses information obtained by solving the lower-dimensional planning problem.

From the construction of the $\delta$-Space, we directly obtain the distance from each state towards the goal along the shortest lower-dimensional path.
We assume that the distance towards the goal is travelled on a straight line and the UAV reaches a zero velocity at the goal.
Thus, we can assume that the UAV accelerates to the maximal reachable velocity, keeps this velocity for a certain time and then decelerates to stop at the goal.
In the following, we describe how this can be incorporated into a search heuristic for second-order trajectories.
The extension to third-order trajectories is left for future work.

For a pair of discretized 1D velocities $v_1, v_2$, let the minimum time needed to accelerate the UAV from $v_1$ to $v_2$ be $t_{v_1 , v_2}$. 
The corresponding control effort is denoted by $c_{v_1 , v_2}$ and the distance the UAV moves during the acceleration by $d_{v_1 , v_2}$.
These values can efficiently be precomputed.
During search, we calculate the maximal velocity that can be reached from an initial velocity $v$ when travelling a distance of $d$ as $v_{max}:=\max\{v' | d_{v , v'} + d_{v', 0} \leq d\}$.
Then, we can estimate the flight time
\begin{equation}
T = \frac{d - d_{v , v_{max}} - d_{v_{max},0}}{v_{max}} + t_{v , v_{max}} + t_{v_{max} , 0}
\end{equation}
and the corresponding control cost
\begin{equation}
c = c_{v,v_{max}} + c_{v_{max},0}.
\end{equation}
For the initial velocity $v$, we choose the maximum of the absolute velocities along the independent axes.
Note that this heuristic is not admissible since it overestimates the cost for diagonal movements.
As our experiments show, this slightly increases the trajectory costs compared to admissible heuristics but it results in lower costs than the inadmissible heuristic used by Liu \etal~\cite{liu_2018_icra}.
 
\section{Evaluation}
\label{sec:Evaluation}
\vspace{-0.2cm}
 \begin{figure}[t]
 \centering 
    \begin{subfigure}{.3\textwidth}
    \centering
    \framebox{\includegraphics[width=.95\textwidth]{img/Liu-path/Complete_1051.jpg}}
    \captionsetup{justification=centering}
    \caption{Full State Space \\ \cite{liu_2018_icra} }
    \label{fig:liu_jerk_experiment:fullstatespace}
    \end{subfigure}
    \begin{subfigure}{.3\textwidth}
    \centering
    \framebox{\includegraphics[width=.95\textwidth]{img/Liu-path/Liu_2_stufig_1051.jpg}}
    \captionsetup{justification=centering}
    \caption{Iterative (Pos. $\rightarrow$ Jerk, $\epsilon = 1.0$) \cite{liu_2018_icra}}
    \label{fig:liu_jerk_experiment:liu_e1.0}
    \end{subfigure}
    \begin{subfigure}{.3\textwidth}
    \centering
    \framebox{\includegraphics[width=.95\textwidth]{img/Liu-path/Liu_e60_1051.jpg}}
    \captionsetup{justification=centering}
    \caption{Iterative (Pos. $\rightarrow$ Jerk, $\epsilon = 0.6$) \cite{liu_2018_icra}}
    \label{fig:liu_jerk_experiment:liu_e0.6}
    \end{subfigure}
    
    \vspace{1em}
    \begin{subfigure}{.3\textwidth}
    \centering
    \framebox{\includegraphics[width=.95\textwidth]{img/Liu-path/Tunnel_1051.jpg}}
    \captionsetup{justification=centering}
    \caption{Tunnel}
    \label{fig:liu_jerk_experiment:tunnel}
    \end{subfigure}
    \begin{subfigure}{.3\textwidth}
    \centering
    \framebox{\includegraphics[width=.95\textwidth]{img/Liu-path/Delta_1051.jpg}}
    \captionsetup{justification=centering}
    \caption{$\delta$-Space}
    \label{fig:liu_jerk_experiment:delta-space}
    \end{subfigure}
    \begin{subfigure}{.3\textwidth}
    \centering
    \framebox{\includegraphics[width=.95\textwidth]{img/Liu-path/DeltaHeuristic_1051.jpg}}
    \captionsetup{justification=centering}
    \caption{$\delta$-Space Heuristic}
    \label{fig:liu_jerk_experiment:delta-heuristic}
    \end{subfigure}
    \caption{Examples for generating third-order 2D trajectories with the method of Liu \etal~\cite{liu_2018_icra}. The start is marked by the blue circle. Obstacles and pruned states are black. Expanded states are grey. (\subref{fig:liu_jerk_experiment:fullstatespace}) Searching the full state space finds the optimal trajectory but is computationally expensive. (\subref{fig:liu_jerk_experiment:liu_e1.0}) The iterative heuristic with a high greediness parameter $\epsilon$ has low planning times but results in a slow trajectory which tightly follows the guiding path. (\subref{fig:liu_jerk_experiment:liu_e0.6}) A lower greediness generates a faster trajectory at the cost of higher planning times, but the obtained solution is still close to the 2D path and thus only locally optimal. (\subref{fig:liu_jerk_experiment:tunnel}) The tunnel also results in a suboptimal trajectory. (\subref{fig:liu_jerk_experiment:delta-space}) $\delta$-Spaces find the optimal trajectory while still having low planning times. (\subref{fig:liu_jerk_experiment:delta-heuristic}) The additional $\delta$-Space heuristic further improves the planning time. }
    \label{fig:liu_jerk_experiment}
\end{figure}

\begin{table}[t]
\caption{Planning statistics for using the $\delta$-Space with the method from Liu \etal~\cite{liu_2018_icra} to generate third-order trajectories in a 2D environment. Planning time, number of expansions and trajectory costs are averaged over the tasks where all methods found a solution. Success denotes the fraction of tasks for which a solution was found within the allowed planning time.}
\begin{center}
\setlength\tabcolsep{0.5em}
\begin{tabular}{l|rrrr}
                                                                           & Success           & Time                      & Expansions                             & Costs \\ \hline 
Full State Space \cite{liu_2018_icra}                                      & $74.0\%$          & \SI{214}{\milli\second}   & \num[group-minimum-digits = 4]{12291}  & \num{121.92} \\
Iterative (Pos. $\rightarrow$ Jerk, $\epsilon = 0.6$) \cite{liu_2018_icra} & $89.7\%$          & \SI{334}{\milli\second}   & \num[group-minimum-digits = 4]{6125}   & \num{180.33} \\
Iterative (Pos. $\rightarrow$ Jerk, $\epsilon = 0.7$) \cite{liu_2018_icra} & $96.5\%$          & \SI{233}{\milli\second}   & \num[group-minimum-digits = 4]{4231}   & \num{204.49} \\
Iterative (Pos. $\rightarrow$ Jerk, $\epsilon = 0.8$) \cite{liu_2018_icra} & $96.3\%$          & \SI{97}{\milli\second}    & \num[group-minimum-digits = 4]{1608}   & \num{223.44} \\
Iterative (Pos. $\rightarrow$ Jerk, $\epsilon = 0.9$) \cite{liu_2018_icra} & $95.7\%$          & \SI{76}{\milli\second}    & \num[group-minimum-digits = 4]{1128}   & \num{247.48} \\
Iterative (Pos. $\rightarrow$ Jerk, $\epsilon = 1.0$) \cite{liu_2018_icra} & $95.2\%$          & \bf\SI{28}{\milli\second} & \bf\num[group-minimum-digits = 4]{260} & \num{267.87} \\
Tunnel ($r=\SI{1.0}{\metre}$)                                              & $94.6\%$          & \SI{50}{\milli\second}    & \num[group-minimum-digits = 4]{3153}   & \num{127.25} \\
Tunnel ($r=\SI{1.5}{\metre}$)                                              & $95.4\%$          & \SI{58}{\milli\second}    & \num[group-minimum-digits = 4]{3586}   & \num{124.12} \\
Tunnel ($r=\SI{2.5}{\metre}$)                                              & $95.4\%$          & \SI{67}{\milli\second}    & \num[group-minimum-digits = 4]{4070}   & \num{123.27} \\
Tunnel ($r=\SI{3.5}{\metre}$)                                              & $94.6\%$          & \SI{72}{\milli\second}    & \num[group-minimum-digits = 4]{4328}   & \num{123.12} \\
$\delta$-Space  ($\delta=\SI{1.0}{\metre}$)                                & $95.9\%$          & \SI{115}{\milli\second}   & \num[group-minimum-digits = 4]{6039}   & \num{122.53} \\
$\delta$-Space  ($\delta=\SI{1.5}{\metre}$)                                & $91.7\%$          & \SI{139}{\milli\second}   & \num[group-minimum-digits = 4]{7175}   & \num{122.46} \\
$\delta$-Space  ($\delta=\SI{2.5}{\metre}$)                                & $84.3\%$          & \SI{182}{\milli\second}   & \num[group-minimum-digits = 4]{9014}   & \num{122.37} \\
$\delta$-Space Heuristic  ($\delta=\SI{1.0}{\metre}$)                      & $\mathbf{97.6\%}$ & \SI{53}{\milli\second}    & \num[group-minimum-digits = 4]{1980}   & \num{117.79} \\
$\delta$-Space Heuristic  ($\delta=\SI{1.5}{\metre}$)                      & $96.1\%$          & \SI{62}{\milli\second}    & \num[group-minimum-digits = 4]{2231}   & \num{117.71} \\
$\delta$-Space Heuristic  ($\delta=\SI{2.5}{\metre}$)                      & $\mathbf{97.6\%}$ & \SI{75}{\milli\second}    & \num[group-minimum-digits = 4]{2484}   & \bf\num{117.53}
\end{tabular}
\end{center}
\label{tab:liu_jerk_experiment}
\end{table}

In a first experiment, we use $\delta$-Spaces on top of the method of Liu \etal~\cite{liu_2018_icra} to plan third-order trajectories in a cluttered 2D environment of size \SI{80}{}$\times$\SI{62}{\meter}. 
We use the source code\footnote{\url{https://github.com/sikang/motion_primitive_library}} from Liu \etal~\cite{liu_2018_icra} with the following parameters:
\begin{center}
\setlength\tabcolsep{0.5em}
\begin{tabular}{c|c|c|c|c|c|c}
$\rho$ & $\tau$            & $v_\text{max}$            & $a_\text{max}$                    & $u_\text{max}$                    & $du$ & Timeout \\ \hline
10     & \SI{1.0}{\second} & \SI{3}{\meter\per\second} &  \SI{1}{\metre\per\square\second}  &  \SI{1}{\metre\per\cubic\second}  &  \SI{0.5}{\metre\per\cubic\second} & \SI{1}{\second}
\end{tabular} .
\end{center}
All other parameters were not changed and are the default ones from the implementation \cite{liu_2018_icra}.
To reduce planning times, Liu \etal propose to generate a lower-order trajectory first, which can then be used as a heuristic to guide the higher-dimensional search.
In this experiment, we choose a 2D path as guiding trajectory for a fair comparison against the $\delta$-Space.
The resulting trajectories for one example task are shown in \reffig{fig:liu_jerk_experiment} and the average planning performance over a set of \num{1000} tasks is summarized in \reftab{tab:liu_jerk_experiment}.
The iterative heuristic of Liu \etal uses a greediness parameter $\epsilon$ to control the deviation of the high-dimensional trajectory from the guiding 2D path.
With a high parameter, the trajectory tightly follows the shortest path (\reffig{fig:liu_jerk_experiment:liu_e1.0}).
This results in planning times that are significantly faster than for all other methods.
However, these trajectories can only be traversed at low velocities and thus, their average costs are more than twice the global optimum.
The costs can be reduced by lowering the greediness parameter, which results in higher planning times.
Note that with a low greediniess of $\epsilon = 0.6$, the method of Liu \etal has significantly higher planning times than the $\delta$-Space method, while still generating trajectories with higher costs.
\reffig{fig:liu_jerk_experiment:liu_e0.6} shows an example trajectory, which is only locally optimal since it still tightly follows the shortest 2D path.
Similar trajectories but higher velocities are obtained when restricting the search to a tunnel around the 2D path (\reffig{fig:liu_jerk_experiment:tunnel}).
Our method (\reffig{fig:liu_jerk_experiment:delta-space}) allows larger deviations from the shortest 2D path and generates a trajectory close to the optimal one.
This result is obtained by using the same (inadmissible) heuristic that Liu \etal use when planning in the full state space (\reffig{fig:liu_jerk_experiment:fullstatespace}).
From the construction of the $\delta$-Space we already know the 2D distances from all states to the goal.
We can use these as a heuristic if we assume that the UAV flies at maximum velocity.
This further reduces the planning times of our method and results in the lowest trajectory costs (\reffig{fig:liu_jerk_experiment:delta-heuristic}).

\begin{figure}[t]
    \centering
    \framebox{\includegraphics[width=0.5\linewidth]{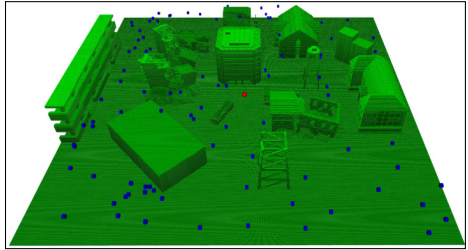}}
    \caption{Simulated 3D outdoor environment. The UAV starts at the map center (red circle) and plans trajectories to all goal positions (blue squares). }
    \label{fig:outdoor_map}
\end{figure} 

\begin{table}[t]
\caption{Planning statistics for using the $\delta$-Space with the method from \cite{schleich_2021_icra} to generate second-order trajectories in a 3D environment. We report results for different expansion schemes during search: standard A* search and a level-based expansion scheme (Level-A*) proposed in \cite{schleich_2021_icra}, which prioritize expansions of states on higher resolution levels. Planning time, number of expansions and trajectory costs are averaged over the tasks where all methods found a solution. Success denotes the fraction of tasks for which a solution was found within the allowed planning time.}
\begin{center}
\setlength\tabcolsep{0.5em}
\begin{tabular}{l|rrrr}
 & Success & Time & Expansions & Costs \\ \hline 
Uniform (A*) & $91.8\%$ & \SI{0.923}{\second}& \num{79883} & \bf\num{251.95} \\
Multiresolution (A*) \cite{schleich_2021_icra} & $95.9\%$ & \SI{1.096}{\second}& \num{81700} & \num{259.55} \\
Tunnel (A*) & $\mathbf{100.0}\%$ & \SI{0.603}{\second}& \num{34994} & \num{259.18} \\
$\delta$-Space (A*) & $98.97\%$ & \SI{0.834}{\second}& \num{55138} & \num{253.05} \\
$\delta$-Space with Heuristic (A*) & $\mathbf{100.0\%}$ & \bf\SI{0.324}{\second}& \bf\num{4695} & \num{264.64} \\ \hline
Uniform (Level-A*) & $97.9\%$ & \SI{0.108}{\second}& \num{9099} & \num{262.82} \\
Multiresolution (Level-A*) \cite{schleich_2021_icra} & $\mathbf{100.0}\%$ & \bf\SI{0.076}{\second}& \num{4491} & \num{277.34} \\
Tunnel (Level-A*) & $\mathbf{100.0}\%$ & \SI{0.287}{\second}& \num{6450} & \num{268.91} \\
$\delta$-Space (Level-A*) & $\mathbf{100.0}\%$ & \SI{0.324}{\second}& \num{6056} & \bf\num{262.14} \\
$\delta$-Space with Heuristic (Level-A*) & $\mathbf{100.0}\%$ & \SI{0.276}{\second}& \bf\num{479} & \num{288.45}
\end{tabular}
\end{center}
\label{tab:mres_experiment}
\end{table}

In a second experiment, we evaluate our method in a simulated outdoor environment (\reffig{fig:outdoor_map}).
Here, we use the method from \cite{schleich_2021_icra} with the same parameters as in the original paper:
\begin{center}
\setlength\tabcolsep{1em}
\begin{tabular}{c|c|c|c|c|c}
$\rho$ & $\tau$ & $v_\text{max}$ & $u_\text{max}$ & $du$ & Timeout\\ \hline
16 & \SI{0.5}{\second} & \SI{4}{\meter\per\second} &  \SI{2}{\metre\per\square\second}  &  \SI{2}{\metre\per\square\second} & \num{1000000} expansions
\end{tabular} .
\end{center}
Note that we abort searches, if no solution within the first \num{1000000} expansions is found.
Although we compare against multiresolutional state lattices, the $\delta$-Space is applied on top of a uniform state lattice, since the $\delta$-Space already reduces the size of the state space.
The results are reported in \reftab{tab:mres_experiment}.
When using standard A* search, the $\delta$-Space outperforms uniform planning in the full state space as well as the multiresolutional planning method from \cite{schleich_2021_icra} with respect to planning time and success rates, while achieving trajectory costs that are close to the optimal ones.
Using the $\delta$-Space heuristic reduces the planning time even further.
However, since the heuristic is not admissible, this comes at the cost of significantly increased path costs.

Note that, compared to uniform planning, multiresolution increases the success rates but has a slightly higher average planning time.
This has already been reported in \cite{schleich_2021_icra}. For some planning tasks, the suboptimal trajectory costs introduced by multiresolution require more expansions during A* search.
To address this issue, a level-based expansion scheme was introduced in \cite{schleich_2021_icra}.
This reduces the planning times at the cost of more suboptimal trajectories.
Here, multiresolution is significantly faster than $\delta$-Space planning.
This is due to the overhead of the low-dimensional search needed to generate the $\delta$-Space, which took on average \SI{0.266}{\second}.
The high-dimensional search itself is actually faster than multiresolutional planning.
Overall, in open-spaced environments, multiresolution can be faster than $\delta$-Space planning but results in higher trajectory costs.
However, $\delta$-Spaces can also be used in more complex environments (like the one in \reffig{fig:liu_jerk_experiment}), where multiresolution is not applicable.

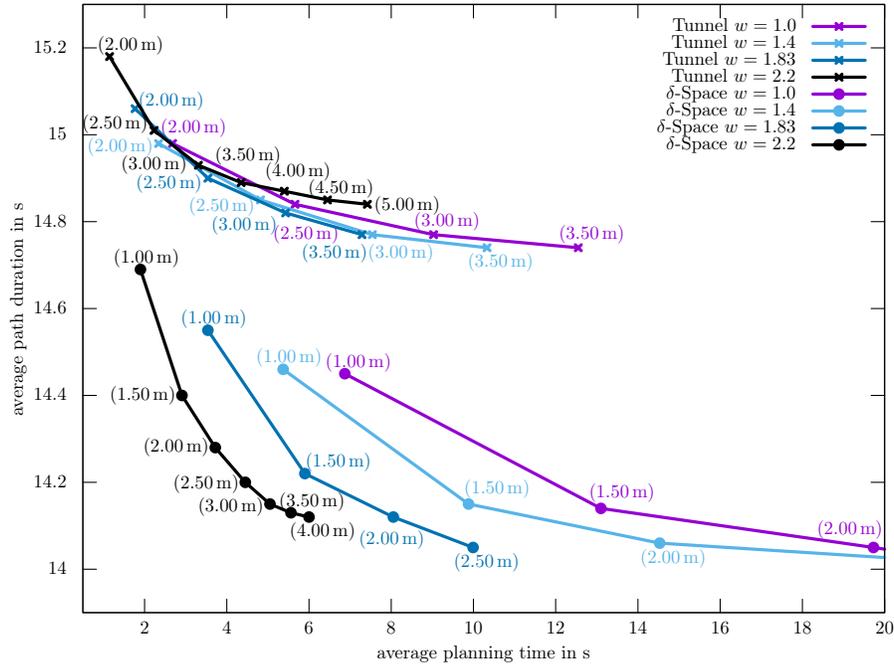
\begin{figure}[t]
\begin{tikzpicture}[gnuplot]
\tikzset{every node/.append style={scale=0.70}}
\path (0.000,0.000) rectangle (12.100,9.000);
\gpcolor{color=gp lt color border}
\gpsetlinetype{gp lt border}
\gpsetdashtype{gp dt solid}
\gpsetlinewidth{1.00}
\draw[gp path] (1.054,1.269)--(1.234,1.269);
\draw[gp path] (11.712,1.269)--(11.532,1.269);
\node[gp node right] at (0.925,1.269) {$14$};
\draw[gp path] (1.054,2.425)--(1.234,2.425);
\draw[gp path] (11.712,2.425)--(11.532,2.425);
\node[gp node right] at (0.925,2.425) {$14.2$};
\draw[gp path] (1.054,3.581)--(1.234,3.581);
\draw[gp path] (11.712,3.581)--(11.532,3.581);
\node[gp node right] at (0.925,3.581) {$14.4$};
\draw[gp path] (1.054,4.737)--(1.234,4.737);
\draw[gp path] (11.712,4.737)--(11.532,4.737);
\node[gp node right] at (0.925,4.737) {$14.6$};
\draw[gp path] (1.054,5.893)--(1.234,5.893);
\draw[gp path] (11.712,5.893)--(11.532,5.893);
\node[gp node right] at (0.925,5.893) {$14.8$};
\draw[gp path] (1.054,7.049)--(1.234,7.049);
\draw[gp path] (11.712,7.049)--(11.532,7.049);
\node[gp node right] at (0.925,7.049) {$15$};
\draw[gp path] (1.054,8.205)--(1.234,8.205);
\draw[gp path] (11.712,8.205)--(11.532,8.205);
\node[gp node right] at (0.925,8.205) {$15.2$};
\draw[gp path] (1.874,0.691)--(1.874,0.871);
\draw[gp path] (1.874,8.783)--(1.874,8.603);
\node[gp node center] at (1.874,0.475) {$2$};
\draw[gp path] (2.967,0.691)--(2.967,0.871);
\draw[gp path] (2.967,8.783)--(2.967,8.603);
\node[gp node center] at (2.967,0.475) {$4$};
\draw[gp path] (4.060,0.691)--(4.060,0.871);
\draw[gp path] (4.060,8.783)--(4.060,8.603);
\node[gp node center] at (4.060,0.475) {$6$};
\draw[gp path] (5.153,0.691)--(5.153,0.871);
\draw[gp path] (5.153,8.783)--(5.153,8.603);
\node[gp node center] at (5.153,0.475) {$8$};
\draw[gp path] (6.246,0.691)--(6.246,0.871);
\draw[gp path] (6.246,8.783)--(6.246,8.603);
\node[gp node center] at (6.246,0.475) {$10$};
\draw[gp path] (7.339,0.691)--(7.339,0.871);
\draw[gp path] (7.339,8.783)--(7.339,8.603);
\node[gp node center] at (7.339,0.475) {$12$};
\draw[gp path] (8.433,0.691)--(8.433,0.871);
\draw[gp path] (8.433,8.783)--(8.433,8.603);
\node[gp node center] at (8.433,0.475) {$14$};
\draw[gp path] (9.526,0.691)--(9.526,0.871);
\draw[gp path] (9.526,8.783)--(9.526,8.603);
\node[gp node center] at (9.526,0.475) {$16$};
\draw[gp path] (10.619,0.691)--(10.619,0.871);
\draw[gp path] (10.619,8.783)--(10.619,8.603);
\node[gp node center] at (10.619,0.475) {$18$};
\draw[gp path] (11.712,0.691)--(11.712,0.871);
\draw[gp path] (11.712,8.783)--(11.712,8.603);
\node[gp node center] at (11.712,0.475) {$20$};
\draw[gp path] (1.054,8.783)--(1.054,0.691)--(11.712,0.691)--(11.712,8.783)--cycle;
\node[gp node center,rotate=-270] at (0.204,4.737) {average path duration in s};
\node[gp node center] at (6.383,0.151) {average planning time in s};
\node[gp node right] at (10.629,8.490) {Tunnel $w=1.0$};
\gpcolor{rgb color={0.580,0.000,0.827}}
\gpsetlinewidth{3.00}
\draw[gp path] (10.758,8.490)--(11.454,8.490);
\draw[gp path] (2.246,6.933)--(3.874,6.124)--(5.716,5.720)--(7.640,5.546);
\gpsetpointsize{4.00}
\gppoint{gp mark 2}{(2.246,6.933)}
\gppoint{gp mark 2}{(3.874,6.124)}
\gppoint{gp mark 2}{(5.716,5.720)}
\gppoint{gp mark 2}{(7.640,5.546)}
\gppoint{gp mark 2}{(11.106,8.490)}
\node[gp node center] at (2.526,7.141) {(\SI{2.00}{\metre})};
\node[gp node center] at (4.02,5.73) {(\SI{2.50}{\metre})};
\node[gp node center] at (5.896,5.898) {(\SI{3.00}{\metre})};
\node[gp node center] at (7.820,5.724) {(\SI{3.50}{\metre})};
\gpcolor{color=gp lt color border}
\node[gp node right] at (10.629,8.265) {Tunnel $w=1.4$};
\gpcolor{rgb color={0.337,0.706,0.914}}
\draw[gp path] (10.758,8.265)--(11.454,8.265);
\draw[gp path] (2.060,6.933)--(3.415,6.182)--(4.902,5.720)--(6.427,5.546);
\gppoint{gp mark 2}{(2.060,6.933)}
\gppoint{gp mark 2}{(3.415,6.182)}
\gppoint{gp mark 2}{(4.902,5.720)}
\gppoint{gp mark 2}{(6.427,5.546)}
\gppoint{gp mark 2}{(11.106,8.265)}
\node[gp node center] at (1.55,6.9) {(\SI{2.00}{\metre})};
\node[gp node center] at (2.9,6.1) {(\SI{2.50}{\metre})};
\node[gp node center] at (5.282,5.478) {(\SI{3.00}{\metre})};
\node[gp node center] at (6.607,5.354) {(\SI{3.50}{\metre})};
\gpcolor{color=gp lt color border}
\node[gp node right] at (10.629,8.040) {Tunnel $w=1.83$};
\gpcolor{rgb color={0.000,0.447,0.698}}
\draw[gp path] (10.758,8.040)--(11.454,8.040);
\draw[gp path] (1.748,7.396)--(2.716,6.471)--(3.749,6.009)--(4.765,5.720);
\gppoint{gp mark 2}{(1.748,7.396)}
\gppoint{gp mark 2}{(2.716,6.471)}
\gppoint{gp mark 2}{(3.749,6.009)}
\gppoint{gp mark 2}{(4.765,5.720)}
\gppoint{gp mark 2}{(11.106,8.040)}
\node[gp node center] at (2.228,7.504) {(\SI{2.00}{\metre})};
\node[gp node center] at (2.2,6.4) {(\SI{2.50}{\metre})};
\node[gp node center] at (3.2,5.85) {(\SI{3.00}{\metre})};
\node[gp node center] at (4.400,5.478) {(\SI{3.50}{\metre})};
\gpcolor{color=gp lt color border}
\node[gp node right] at (10.629,7.815) {Tunnel $w=2.2$};
\gpcolor{rgb color={0.000,0.000,0.000}}
\draw[gp path] (10.758,7.815)--(11.454,7.815);
\draw[gp path] (1.409,8.089)--(2.000,7.107)--(2.590,6.644)--(3.158,6.413)--(3.732,6.298)%
  --(4.306,6.182)--(4.836,6.124);
\gppoint{gp mark 2}{(1.409,8.089)}
\gppoint{gp mark 2}{(2.000,7.107)}
\gppoint{gp mark 2}{(2.590,6.644)}
\gppoint{gp mark 2}{(3.158,6.413)}
\gppoint{gp mark 2}{(3.732,6.298)}
\gppoint{gp mark 2}{(4.306,6.182)}
\gppoint{gp mark 2}{(4.836,6.124)}
\gppoint{gp mark 2}{(11.106,7.815)}
\node[gp node center] at (1.689,8.247) {(\SI{2.00}{\metre})};
\node[gp node center] at (1.48,7.2) {(\SI{2.50}{\metre})};
\node[gp node center] at (2.00,6.652) {(\SI{3.00}{\metre})};
\node[gp node center] at (3.25,6.78) {(\SI{3.50}{\metre})};
\node[gp node center] at (3.912,6.56) {(\SI{4.00}{\metre})};
\node[gp node center] at (4.486,6.340) {(\SI{4.50}{\metre})};
\node[gp node center] at (5.366,6.12) {(\SI{5.00}{\metre})};
\gpcolor{color=gp lt color border}
\node[gp node right] at (10.629,7.590) {$\delta$-Space $w=1.0$};
\gpcolor{rgb color={0.580,0.000,0.827}}
\draw[gp path] (10.758,7.590)--(11.454,7.590);
\draw[gp path] (4.536,3.870)--(7.941,2.078)--(11.564,1.558)--(11.712,1.536);
\gppoint{gp mark 7}{(4.536,3.870)}
\gppoint{gp mark 7}{(7.941,2.078)}
\gppoint{gp mark 7}{(11.564,1.558)}
\gppoint{gp mark 7}{(11.106,7.590)}
\node[gp node center] at (4.716,4.028) {(\SI{1.00}{\metre})};
\node[gp node center] at (8.221,2.286) {(\SI{1.50}{\metre})};
\node[gp node center] at (11.25,1.8) {(\SI{2.00}{\metre})};
\gpcolor{color=gp lt color border}
\node[gp node right] at (10.629,7.365) {$\delta$-Space $w=1.4$};
\gpcolor{rgb color={0.337,0.706,0.914}}
\draw[gp path] (10.758,7.365)--(11.454,7.365);
\draw[gp path] (3.716,3.928)--(6.181,2.136)--(8.722,1.616)--(11.712,1.424);
\gppoint{gp mark 7}{(3.716,3.928)}
\gppoint{gp mark 7}{(6.181,2.136)}
\gppoint{gp mark 7}{(8.722,1.616)}
\gppoint{gp mark 7}{(11.106,7.365)}
\node[gp node center] at (3.796,4.086) {(\SI{1.00}{\metre})};
\node[gp node center] at (6.561,2.35) {(\SI{1.50}{\metre})};
\node[gp node center] at (8.902,1.424) {(\SI{2.00}{\metre})};
\gpcolor{color=gp lt color border}
\node[gp node right] at (10.629,7.140) {$\delta$-Space $w=1.83$};
\gpcolor{rgb color={0.000,0.447,0.698}}
\draw[gp path] (10.758,7.140)--(11.454,7.140);
\draw[gp path] (2.716,4.448)--(4.005,2.541)--(5.181,1.963)--(6.241,1.558);
\gppoint{gp mark 7}{(2.716,4.448)}
\gppoint{gp mark 7}{(4.005,2.541)}
\gppoint{gp mark 7}{(5.181,1.963)}
\gppoint{gp mark 7}{(6.241,1.558)}
\gppoint{gp mark 7}{(11.106,7.140)}
\node[gp node center] at (2.796,4.606) {(\SI{1.00}{\metre})};
\node[gp node center] at (4.45,2.7) {(\SI{1.50}{\metre})};
\node[gp node center] at (5.161,1.671) {(\SI{2.00}{\metre})};
\node[gp node center] at (6.421,1.366) {(\SI{2.50}{\metre})};
\gpcolor{color=gp lt color border}
\node[gp node right] at (10.629,6.915) {$\delta$-Space $w=2.2$};
\gpcolor{rgb color={0.000,0.000,0.000}}
\draw[gp path] (10.758,6.915)--(11.454,6.915);
\draw[gp path] (1.819,5.257)--(2.371,3.581)--(2.814,2.887)--(3.213,2.425)--(3.541,2.136)%
  --(3.820,2.020)--(4.060,1.963);
\gppoint{gp mark 7}{(1.819,5.257)}
\gppoint{gp mark 7}{(2.371,3.581)}
\gppoint{gp mark 7}{(2.814,2.887)}
\gppoint{gp mark 7}{(3.213,2.425)}
\gppoint{gp mark 7}{(3.541,2.136)}
\gppoint{gp mark 7}{(3.820,2.020)}
\gppoint{gp mark 7}{(4.060,1.963)}
\gppoint{gp mark 7}{(11.106,6.915)}
\node[gp node center] at (1.899,5.415) {(\SI{1.00}{\metre})};
\node[gp node center] at (1.851,3.589) {(\SI{1.50}{\metre})};
\node[gp node center] at (2.294,2.895) {(\SI{2.00}{\metre})};
\node[gp node center] at (2.693,2.413) {(\SI{2.50}{\metre})};
\node[gp node center] at (3.021,2.104) {(\SI{3.00}{\metre})};
\node[gp node center] at (4.10,2.178) {(\SI{3.50}{\metre})};
\node[gp node center] at (4.240,1.771) {(\SI{4.00}{\metre})};
\gpcolor{color=gp lt color border}
\gpsetlinewidth{1.00}
\draw[gp path] (1.054,8.783)--(1.054,0.691)--(11.712,0.691)--(11.712,8.783)--cycle;
\gpdefrectangularnode{gp plot 1}{\pgfpoint{1.054cm}{0.691cm}}{\pgfpoint{11.712cm}{8.783cm}}
\end{tikzpicture}
     \caption{The impact of weighted heuristics and anytime planning. A line represents an iterative search as described in \refsec{sec:iterative_delta}.}
    \label{exp:3D:different_heuristics:map1}
\end{figure}

Finally, we evaluate the anytime version of our planning algorithm.
It iteratively increases the size of the $\delta$-Space but can also be applied to plan in tunnels of increasing radius around the low-dimensional path.
In this experiment, we investigate which state space representation is more suitable for the anytime planning algorithm.
Furthermore, we analyse the influence of different heuristic weights that might speed up planning at the cost of sub-optimal solutions.
Since the anytime algorithm requires significant adjustments of the source code of the high-dimensional planning algorithms, we implemented our own planning method for this experiment.
Our planner generates third-order trajectories on a 3D map.
For this, it uses motion primitives that were precomputed by the time-optimal trajectory generation method from \cite{beul_2017_icuas}.
Figure \ref{exp:3D:different_heuristics:map1} shows the results for both the tunnel and the $\delta$-Space with different heuristic weights. 
The colors of the graphs represents the used weight and a line represents an iterative search as described in \refsec{sec:iterative_delta}.
Here, it can be clearly seen that tunnel and $\delta$-Space behave differently. 
The heuristic weight $w$ does impact the tunnel only slightly.
Increasing the tunnel radius does not improve the trajectory costs much and the costs start to stagnate after few iterations.
On the other hand, the heuristic weight strongly influences the performance of the $\delta$-Spaces. 
A larger weight significantly decreases the planning times.
Although the trajectory costs are increased, this can be made up for by iteratively increasing the size of the $\delta$-Space.
Thus, using the $\delta$-Space with a large heuristic weight generates trajectories with similar planning times as the tunnel method but significantly lower costs.

\begin{table}[t]
\caption{Planning statistics for the anytime algorithm to generate third-order trajectories in a 3D environment.}    
\begin{center}
\begin{tabular}{c r|r|r|r|r|r}
    & & \makecell{iterative \\ planning \\ time}   & \makecell{accumulated \\ iterative \\ planning \\ time} & \makecell{direct \\ planning \\ time}  & \makecell{trajectory \\ cost \\ (iterative)}  & \makecell{ trajectory \\ cost \\ (direct)}   \\ \hline
\multirow{4}*{\makecell{$\delta$-Space }}               & $\delta=$\SI{1.0}{\metre}     & $+3.5s$  & $3.5s$  & $3.5s$ & $14.55s$ & $14.55s$ \\
                                                        & $\delta=$\SI{1.5}{\metre}     & $+2.4s$ & $5.9s$  & $4.7s$ & $14.22s$ & $14.25s$ \\
                                                        & $\delta=$\SI{2.0}{\metre}     & $+2.2s$ & $8.1s$  & $6.3s$ & $14.12s$ & $14.17s$ \\
                                                        & $\delta=$\SI{2.5}{\metre}     & $+1.9s$ & $10.0s$ & $7.7s$ & $14.05s$ & $14.09s$ \\ \hline
\multirow{4}*{\makecell{Tunnel}}                        & $r=$\SI{1.0}{\metre}          & $+0.6s$ & $0.6s$  & $0.6s$ & $16.01s$ & $16.01s$\\
                                                        & $r=$\SI{1.5}{\metre}          &$+0.9s$  & $1.5s$  & $1.2s$ & $15.27s$ & $15.25s$ \\
                                                        & $r=$\SI{2.0}{\metre}          &$+1.3s$  & $2.8s$  & $1.8s$ & $15.03s$ & $15.06s$ \\
                                                        & $r=$\SI{2.5}{\metre}          &$+1.8s$  & $4.6s$  & $2.8s$ & $14.90s$ & $14.94s$
\end{tabular}

\end{center}
\label{tab:iterative_delta}
\end{table}

\reftab{tab:iterative_delta} summarizes the planning performance for each iteration of the anytime algorithms using a heuristic weight $w=1.83$.
As expected, the accumulated planning time for iterative planning is higher than the time for directly planning in the corresponding state space.
However, the relative increase is smaller for the $\delta$-Space, compared to the Tunnel.
This shows that $\delta$-Spaces are more suitable for anytime planning.
Interestingly, the trajectory costs are slightly lower for anytime planning.
Since each iteration allows additional re-expansions, iteratively increasing $\delta$ can help to correct trajectories that are suboptimal due to the inadmissible heuristic.
 
\section{Conclusion}
\label{sec:Conclusion}
\vspace{-0.2cm}
In this paper, we propose a new method for pruning of high-dimensional state spaces.
Instead of restricting the search to the vicinity of an optimal lower-dimensional path, we introduce the $\delta$-Space, which represents the union of multiple paths within a suboptimality bound $\delta$.
Our experiments show that this significantly reduces planning times.
In contrast to other state space reduction methods, $\delta$-Spaces are less prone to local minima and can find trajectories close to the globally optimal solution.
Furthermore, we introduced an anytime planning algorithm that efficiently plans in $\delta$-Spaces of increasing sizes.
Thus, a first initial solution can be found quickly, while subsequent iterations further improve the solution quality.

\vspace*{1ex}
{\footnotesize\noindent\textbf{Acknowledgements.} This work has been funded by the German Federal Ministry of Education and Research (BMBF) in the project ``Kompetenzzentrum: Aufbau des Deutschen Rettungsrobotik-Zentrums (A-DRZ)''.}

\bibliography{literature_references}{}
\bibliographystyle{splncs}

\end{document}